\documentclass{article}

\usepackage{microtype}
\usepackage{graphicx}
\usepackage{subcaption}
\usepackage{booktabs} 

\usepackage{hyperref}

\usepackage[accepted]{icml2026}

\usepackage{amsmath}
\usepackage{amssymb}
\usepackage{mathtools}
\usepackage{amsthm}

\usepackage[capitalize,noabbrev]{cleveref}

\newcommand{\rvx}{\mathbf{x}}

\newcommand{\vx}{\boldsymbol{x}}

\newcommand{\vu}{\boldsymbol{u}}

\newcommand{\veps}{\boldsymbol{\epsilon}}

\newcommand{\vmu}{\boldsymbol{\mu}}

\theoremstyle{plain}

\theoremstyle{definition}

\theoremstyle{remark}

\renewcommand{\algorithmiccomment}[1]{\hfill$\triangleright$ #1}

\usepackage[textsize=tiny]{todonotes}

\icmltitlerunning{Uncertainty Estimation for Molecular Diffusion Models}

\begin{document}

\twocolumn[
  \icmltitle{Uncertainty Estimation for Molecular Diffusion Models}



  \icmlsetsymbol{equal}{*}

   \begin{icmlauthorlist}
    \icmlauthor{Paul Seij}{delta}
    \icmlauthor{Christian A. Naesseth}{delta}
    \icmlauthor{Stephan Mandt}{uci}
    \icmlauthor{Metod Jazbec}{delta}
  \end{icmlauthorlist}

  \icmlaffiliation{delta}{UvA-Bosch Delta Lab, University of Amsterdam}
  \icmlaffiliation{uci}{University of California, Irvine}

  \icmlcorrespondingauthor{Metod Jazbec}{m.jazbec@uva.nl}

  \icmlkeywords{Machine Learning, ICML}

  \vskip 0.3in
]



\printAffiliationsAndNotice{}  

\begin{abstract}
 Diffusion models have seen wide adoption for 3D molecular generation, yet they offer no principled signal of when a generated molecule is likely to be of low quality. We propose a post-hoc method for estimating per-sample uncertainty in pretrained molecular diffusion models. Building on a Laplace approximation of the denoising network, we measure the variability of the noise prediction across the generation trajectory. Empirically, we show that the resulting uncertainty score is informative of sample quality, exhibiting a negative correlation with established sample-level quality metrics. We further study how the proposed uncertainty score can be used to filter generated samples, improving model performance via test-time scaling.
\end{abstract}

\section{Introduction}
Diffusion models have rapidly become the dominant paradigm for 3D molecular generation \citep{hoogeboom2022equivariant,xu2023geometric,morehead2024geometry,cornet2024equivariant}. Yet even state-of-the-art models can produce samples that are chemically invalid or unstable. This becomes a practical bottleneck when scaling up the generation budgets: deciding which molecules merit expensive downstream evaluation (e.g., docking or wet-lab validation) currently relies on coarse rule-based filters or on the model's own---often poorly calibrated \citep{theis2015note,nalisnick2018deep}---likelihood.

For predictive models, uncertainty estimation is the standard tool for flagging unreliable outputs \citep{lakshminarayanan2017simple}, and a growing line of work has begun to adapt these ideas to diffusion models in the image domain \citep{kou2023bayesdiff,berry2024shedding,aithal2024understanding,jazbec2025generative,de2025diffusion}, showing promising early results for flagging low-quality generated images. To the best of our knowledge, however, no analogous work exists for molecular diffusion, despite the aforementioned practical need for sample-quality signals in scientific domains where downstream evaluation is expensive.

In this work, we take a first step towards extending the existing diffusion uncertainty approaches to molecular domain. Our method is post-hoc: given a pretrained molecular diffusion model, we fit a Laplace approximation \citep{daxberger2021laplace} to the denoising network and measure the variability of its noise predictions along the generation trajectory. We aggregate this variability across selected timesteps, atoms, and feature dimensions to obtain a single uncertainty score for each generated molecule. Empirically, we show that this score is informative of molecular sample quality: it is negatively correlated with established metrics such as molecular stability, atom stability, and validity, and it is more predictive than diffusion likelihood baseline in our QM9 \citep{ramakrishnan2014quantum} experiments. We further show that the score can be used for test-time scaling \citep{ma2025inference,lee2025adaptive} by oversampling molecules and filtering out high-uncertainty generations, improving sample quality without retraining the underlying generator.

\section{Background} 

\paragraph{Diffusion Models} Diffusion models \citep{sohl2015deep,song2020score,karras2022elucidating} are a class of generative models that learn to invert a gradual noising process. Given a data sample $\vx_0 \sim q(\rvx)$, the forward process produces progressively noisier samples via
\begin{align}
\label{eq:forward}
\vx_t = \sqrt{\bar{\alpha}_t}\, \vx_0 + \sqrt{1 - \bar{\alpha}_t}\, \veps, \qquad \veps \sim \mathcal{N}(0, I),
\end{align}
where $\bar{\alpha}_t = \prod_{r=1}^t (1 - \beta_r)$ and $\{\beta_r\}_{r=1}^T$ is a variance schedule with $\beta_r \in (0,1)$, chosen such that $\vx_T$ is approximately standard Gaussian. For any $s < t$, the reverse-time posterior is also Gaussian distributed  
\begin{align*}
q(\vx_{s} \mid \vx_t, \vx_0) = \mathcal{N}\!\left(\vx_{s}; \vmu_{s\rightarrow t}(\vx_t, \vx_0),\, \beta_{s \rightarrow t} I\right),
\end{align*}
where $\vmu_{s\rightarrow t}$ and $\beta_{s \rightarrow t}$ can be obtained analytically \citep{ho2020denoising}. Since $\vx_0$ is unknown at generation time, the learned reverse process $p_\theta(\vx_{s} \mid \vx_t)$ is trained to match this posterior. In practice, the mean is often reparametrized in terms of the noise ($\veps$) prediction which yields the simplified regression objective \citep{ho2020denoising}\footnote{Among other possible parameterizations of the denoising target; see \citet{song2020score, karras2022elucidating}.}
\begin{align}
\label{eq:diff-loss}
\mathcal{L}(\theta) = \mathbb{E}_{\vx_0,\veps,t} \left[ \|\veps - \hat{\veps}_t\|_2^2 \right],
\end{align}
with $\hat{\veps}_t := f_\theta(\vx_t, t)$. After training, starting from $\hat{\vx}_T \sim \mathcal{N}(0, I)$ a new sample $\hat{\vx}$ is generated by iteratively predicting noise $\hat{\veps}_t$ and sampling a partially denoised $\hat{\vx}_{t-1} \sim p_{\hat{\theta}}(\cdot \mid \hat{\vx}_t, \hat{\veps}_t)$ along a decreasing sequence of timesteps $t=T, T-1,\ldots, 1$, according to the transition rules of a chosen sampler, such as DDPM \citep{ho2020denoising} or DDIM \citep{song2020denoising}. 

Beyond sampling, diffusion models also admit tractable likelihood evaluation. In the discrete-time setting we consider, an estimate of $\log p_{\hat\theta}(\vx)$ for any sample $\vx$ is obtained by evaluating a variational lower bound (ELBO) \citep{ho2020denoising,kingma2021variational}; in the continuous-time formulation, the analogous quantity follows from solving the probability flow ODE \citep{song2020score}.

\section{Methods}
\label{sec:methods}

We next describe our approach for estimating the uncertainty of molecules generated by diffusion models. The goal is to produce a single uncertainty score $u \in \mathbb{R}_{\ge 0}$ that correlates with the quality of a generated sample $\hat{\vx} \in \mathbb{R}^{N_a \times D}$, where $N_a$ is the number of atoms and $D$ the number of per-atom features (including coordinates). At a high level, we measure the variability of the denoiser's prediction across sampling steps, based on intuition that samples the model is internally uncertain about should exhibit more variable denoising trajectories, and aggregate this variability over time, atoms, and features into a single score (see \Cref{alg:mol_unc}).

\begin{algorithm}[t]
\caption{Molecular Generation with Uncertainty}
\label{alg:mol_unc}
\begin{algorithmic}[1]
\STATE \textbf{Input:} diffusion model $f_{\hat{\theta}}$, Laplace posterior $q(\theta) = \mathcal{N}(\hat{\theta}, \Sigma )$, uncertainty time-steps $\mathcal{T}$
\STATE \textbf{Output:} molecule $\hat{\vx} \in \mathbb{R}^{N_a \times D}$, uncertainty $u \in \mathbb{R}_{\ge0}$
\STATE $\hat{\vx}_T \sim \mathcal{N}(0, I) \in \mathbb{R}^{N_a \times d}$
\FOR{$t=T,\ldots,1$}
\STATE $\hat{\veps}_t = f_{\hat{\theta}}(\hat{\vx}_t, t)$ 
\STATE $\hat{\vx}_{t-1} \sim p_{\hat{\theta}}(\cdot \mid \hat{\vx}_t, \hat{\veps}_t )$ \COMMENT{sample (e.g., DDPM)}
\IF{$t \in \mathcal{T}$}
\FOR{$m=1,\ldots,M$}
\STATE $\veps_t^m = f_{\theta_m}(\hat{\vx}_t, t), \quad \theta_m \sim q(\theta)$
\ENDFOR
\STATE $\bar{\veps}_t = \frac{1}{M}\sum_{m=1}^M \veps_t^m$
\STATE $\vu_t = \frac{1}{M}\sum_{m=1}^M (\veps_t^m - \bar{\veps}_t)^2$
\ENDIF
\ENDFOR
\STATE $u = \frac{1}{|\mathcal{T}| \cdot N_a \cdot D}  \sum_{t \in \mathcal{T}} \sum_{n=1}^{N_a}\sum_{d=1}^D [\vu_t]_{n,d}$
\end{algorithmic}
\end{algorithm}

\paragraph{Laplace Posterior} As a first step, we fit an approximate posterior over model parameters, $q(\theta) \approx p(\theta \mid \mathcal{D})$. To remain post-hoc compatible with any pretrained molecular diffusion model, we follow prior work on uncertainty estimation in diffusion models \citep{kou2023bayesdiff, jazbec2025generative} and use the Laplace approximation \citep{mackay1992practical},
\begin{align}
\label{eq:la}
q(\theta) = \mathcal{N}(\theta \mid \hat{\theta}, \Sigma), \quad \Sigma = \big(\nabla^2_{\theta} \mathcal{L}(\theta; \mathcal{D}) \big|_{\hat{\theta}} + \lambda I\big)^{-1},
\end{align}
where $\hat{\theta}$ are the parameters of the pretrained diffusion model, $\mathcal{L}$ is the diffusion training loss from Eq.~\ref{eq:diff-loss}, and $\lambda$ is the prior precision. To scale to models with many parameters, we follow \citet{daxberger2021laplace} and apply probabilistic treatment only to the final layers of the network as well as use the empirical Fisher as a Hessian approximation.

\paragraph{Noise Prediction Variability} Next, we use the posterior $q(\theta)$ to measure the variability of the denoiser $f_{\theta}$ along a sampling trajectory $\{\hat{\vx}_t\}_{t=T,\ldots,0}$. Concretely, for a subset of sampling steps $\mathcal{T} \subseteq \{0, \ldots, T\}$, we draw $M$ weight samples from the posterior and compute the corresponding noise predictions,
\begin{align*}
    \veps_t^m = f_{\theta_m}(\hat{\vx}_t, t), \quad \theta_m \sim q(\theta),
\end{align*}
and define the intermediate uncertainty $\vu_t \in \mathbb{R}_{\ge 0}^{N_a \times D}$ as their elementwise sample variance,
\begin{align*}
    \vu_t = \frac{1}{M}\sum_{m=1}^M (\veps_t^m - \bar{\veps}_t)^2, \quad \bar{\veps}_t = \frac{1}{M}\sum_{m=1}^M \veps_t^m.
\end{align*}
Empirically, we find that a small subset of sampling steps ($|\mathcal{T}| \ll T$) is sufficient for good performance, which limits the computational overhead of uncertainty estimation. Finally, we aggregate the per-step uncertainties $\{\vu_t\}_{t \in \mathcal{T}}$ into a single score $u (\hat{\vx})$ for the generated sample $\hat{\vx}$ by averaging across time steps, atoms, and feature dimensions:
\begin{align}
\label{eq:unc_score}
    u(\hat{\vx}) := \frac{1}{|\mathcal{T}| \cdot N_a \cdot D} \sum_{t \in \mathcal{T}} \sum_{n=1}^{N_a} \sum_{d=1}^{D} [\vu_t]_{n,d}.
\end{align}

\begin{figure*}[t]
\centering
\includegraphics[width=0.99\linewidth]{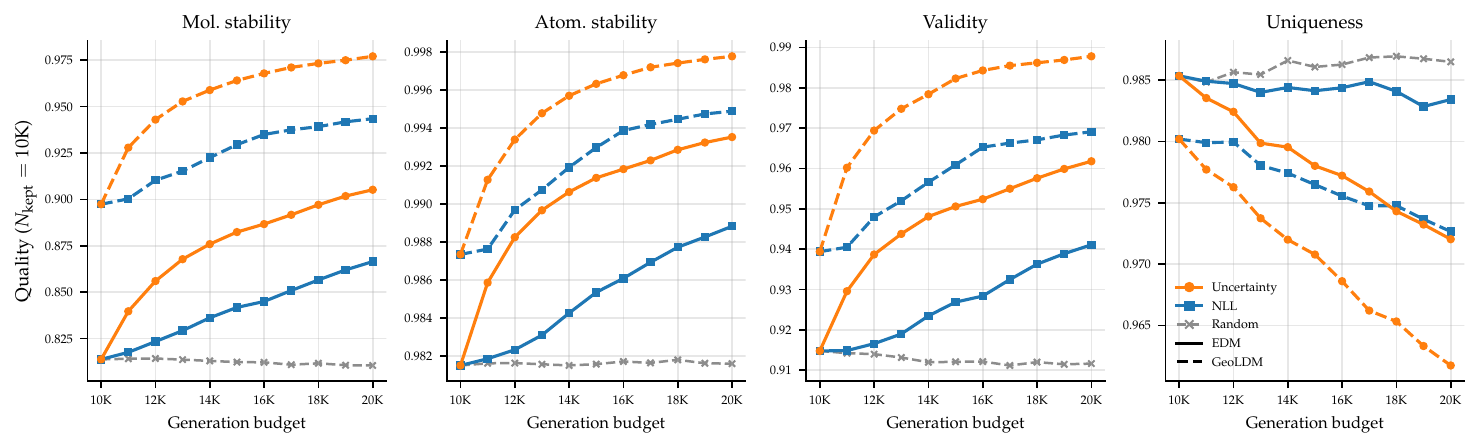}
\caption{Sample quality vs.\ generation budget on QM9 dataset under filtering-based test-time scaling for EDM and GeoLDM diffusion models. We generate $N \in \{10\text{K}, 11\text{K}, \ldots, 20\text{K}\} $ molecules  and keep the $10$K with the lowest uncertainty or lowest diffusion NLL ; random subsampling and the no-filter setting at $N=10$K serve as references. Uncertainty-based filtering improves molecular stability, atomic stability, and validity on both models and consistently outperforms NLL-based filtering, at the cost of a ${\sim}1\%$ drop in uniqueness. Notably, the EDM gains from test-time filtering are comparable in magnitude to those from switching the underlying model to GeoLDM at the $10$K generation budget.}
\label{fig:main_qm9}
\end{figure*}

\section{Experiments}
In our experiments, we first establish that our proposed uncertainty score (Equation \ref{eq:unc_score}) is informative of sample quality by reporting its correlation with standard sample-level metrics used in molecular generation (\Cref{sec:exp_corr}). Next, we explore how uncertainty can be used in a test-time compute setting \citep{ma2025inference}, where we oversample generated molecules and use uncertainty as a filtering criterion (\Cref{sec:exp_ttscale}). Finally, we report a series of ablations to support our proposed design choices (\Cref{sec:exp_abl}).

\paragraph{Experimental details} We evaluate our approach on two open-source diffusion models for 3D molecular generation: i) EDM \citep{hoogeboom2022equivariant} is an E(3)-equivariant diffusion model that jointly diffuses atomic coordinates and features in data space, while ii) GeoLDM \citep{xu2023geometric} performs diffusion in a learned equivariant latent space. For both models, we use the official pretrained checkpoints, as our uncertainty estimation is post-hoc and does not require retraining. We use two standard datasets: QM9 \citep{ramakrishnan2014quantum}, containing small organic molecules with up to 9 heavy atoms, and GEOM-Drugs \citep{axelrod2022geom}, a larger collection of drug-like molecules with substantially more atoms and conformational complexity. For the Laplace approximation, we use default hyperparameters (prior precision $\lambda=1$, likelihood noise $\sigma=1$). We measure uncertainty only in the final $10\%$ of generation process, i.e. $t<100$ when using $T=1000$, as we found that including earlier timesteps yields no or only marginal performance improvements (\Cref{sec:exp_abl}).

\subsection{Uncertainty correlates with sample quality} \label{sec:exp_corr} To assess whether our proposed uncertainty is predictive of sample quality, we report its correlation with a set of well-established metrics for molecular generation in \Cref{tab:main_results} on QM9 dataset. Across both models, our uncertainty score exhibits a statistically-significant negative correlation, indicating that the measured variability is picking up signal informative of sample quality. Moreover, uncertainty exhibits a stronger negative correlation compared to using diffusion model log-likelihood $\log p_{\hat{\theta}}(\hat{\vx})$ (NLL). For example, using the EDM model uncertainty achieves $-0.284$ correlation with molecular stability, while likelihood gets $-0.15$.  

\begin{table}[t]
\centering
\caption{Comparison of uncertainty estimation approaches across model-dataset pairs. We report Spearman rank correlation between the uncertainty score and each quality metric (lower is better). For \emph{ours} see \Cref{alg:mol_unc}, likelihood is based on (negative )diffusion log-likelihood $\log p_{\hat{\theta}}(\hat{\vx})$. For all model/data settings, we report results based on $N{=}10$K generated molecules.} 
\label{tab:main_results}
\small                                                                              
  \begin{tabular}{l cc cc}                                                         
  \toprule                                                                                                                
   & \multicolumn{2}{c}{EDM / QM9} & \multicolumn{2}{c}{GeoLDM / QM9} \\                                                                                          
  \cmidrule(lr){2-3} \cmidrule(lr){4-5}                                           
  Metric & Uncertainty & NLL & Uncertainty & NLL \\                               
  \midrule                                                                          
  Mol.\ stab.  & $-$0.284 & $-$0.150 & $-$0.333 & $-$0.171 \\                    
  Atom.\ stab. & $-$0.305 & $-$0.160 & $-$0.334 & $-$0.178 \\                                                    
  Validity         & $-$0.212 & $-$0.100 & $-$0.266 & $-$0.147 \\                     
  \bottomrule                                
  \end{tabular} 
  
  \end{table}  

\subsection{Test-time scaling via uncertainty}
\label{sec:exp_ttscale}
We next examine whether the observed correlation between our uncertainty score and molecular sample quality can be exploited in practice. Following a test-time scaling setting \citep{ma2025inference}, we improve the quality of a fixed-size set of generated molecules via oversampling and filtering: we generate $N$ samples, with $N$ varying from $10$K to $20$K, and retain the $10$K with the lowest estimated uncertainty (or, as a baseline, the lowest diffusion NLL). The $N{=}10$K case corresponds to no filtering.

\begin{figure*}[t]
\centering
\includegraphics[width=0.99\linewidth]{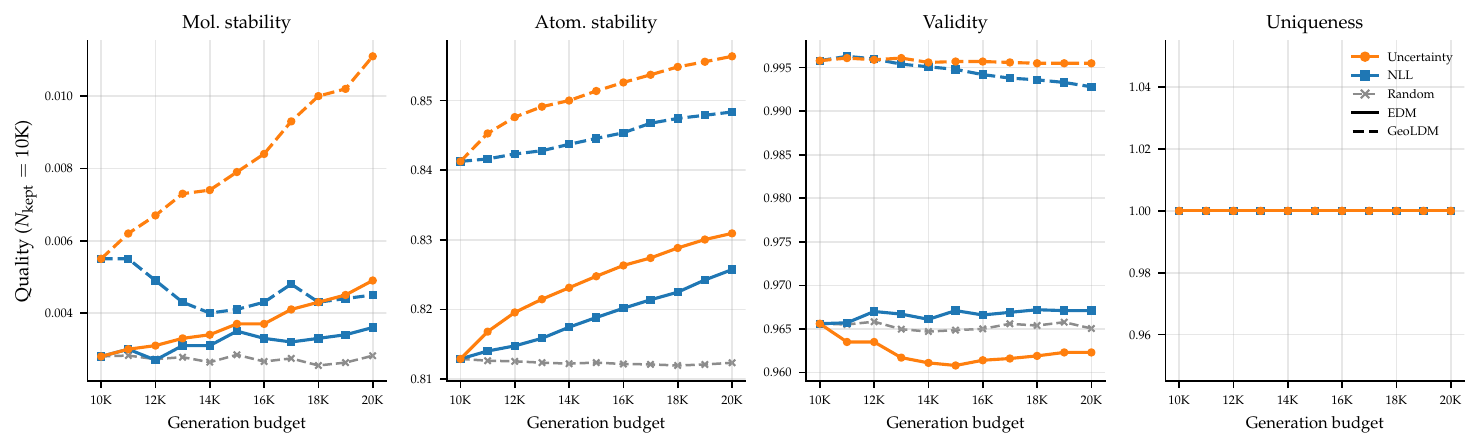}
  \caption{Same setup as Figure~\ref{fig:main_qm9}, evaluated on GEOM-Drugs. In contrast to QM9, neither uncertainty- nor NLL-based filtering meaningfully improves molecular stability, atom stability, or validity over random subsampling on this larger, more complex dataset. }
\label{fig:main_drugs}
\end{figure*}

Focusing first on QM9 (\Cref{fig:main_qm9}), we find that uncertainty-based filtering meaningfully improves molecular stability, atomic stability, and validity across both EDM and GeoLDM. For EDM, oversampling to $20$K and filtering down to $10$K yields gains of ${\sim}10\%$ in molecular stability, ${\sim}1\%$ in atomic stability, and ${\sim}5\%$ in validity. Notably, these test-time gains are comparable in magnitude to the improvement from switching the underlying model from EDM to GeoLDM at the $10$K generation budget, highlighting test-time compute as a complementary axis for improving molecular generation. The gains come with a modest tradeoff in diversity: uniqueness drops by roughly $1\%$. Finally, the stronger negative correlation of uncertainty relative to NLL translates into larger downstream gains: uncertainty-based filtering outperforms NLL-based filtering across both models and all three quality metrics (molecular stability, atomic stability, validity) on QM9.

On GEOM-Drugs (\Cref{fig:main_drugs}), by contrast, test-time filtering fails to meaningfully improve any of the three sample quality metrics, both for our uncertainty score and for NLL. We leave an investigation of why uncertainty-based filtering does not transfer to this larger, more complex dataset to future work.

\begin{figure}[t]
\centering
\includegraphics[width=\linewidth]{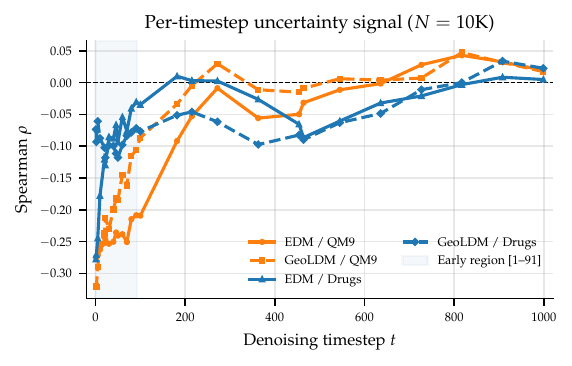}
\caption{Spearman rank correlation between single-timestep score uncertainty $u_t$ and atomic stability, as a function of timestep $t$. Signal concentrates sharply at the clean end of the generation trajectory ($t \approx 91$) for both EDM and GeoLDM; measuring uncertainty at the start of the generation (large $t$) provides weaker signal.}
\label{fig:step_profile}
\end{figure}

\subsection{Ablations}
\label{sec:exp_abl}

We begin our ablations by isolating the contribution of the Laplace approximation (Equation \ref{eq:la}). We compare the quality signal from our uncertainty score against a simpler alternative that omits the Fisher approximation of the Hessian and instead samples from an isotropic Gaussian centered at the MAP \citep{zhdanov2025identity}:
\begin{align}
\label{eq:laplace_ablate}
    q_{\lambda}(\theta) = \mathcal{N}(\theta \mid \hat{\theta}, \lambda^{-1} I) \: .
\end{align}
As shown in \Cref{tab:laplace_ablation}, the simpler, Fisher-free variant performs almost on par with the original Laplace approximation in terms of yielding uncertainties that correlate (negatively) with the quality metrics like molecular stability. This suggests that the signal driving our score is closer to a sensitivity measure --- the variability of model predictions under small isotropic perturbations of the MAP --- than to `true' epistemic uncertainty in a strict Bayesian sense.

\begin{table}[t]
\centering
 \caption{Ablation on the Laplace approximation (EDM / QM9, $N=10$K). We compare our full Fisher-based Laplace posterior against a Fisher-free variant that replaces the covariance with an isotropic Gaussian (Equation \ref{eq:laplace_ablate}). Same as in \Cref{tab:main_results}, we report  Spearman rank correlations between the uncertainty score and each quality metric. The two variants perform almost on par, suggesting our score behaves more like a sensitivity measure than `Bayesian' epistemic uncertainty.}
\label{tab:laplace_ablation}
\small
\begin{tabular}{l cc}
\toprule
   & \multicolumn{2}{c}{EDM / QM9} \\
   \cmidrule(lr){2-3}                
  Metric & Ours  & Ours (w/o Fisher)  \\
  \midrule                                                           
  Mol.\ stab.  & $-$0.284 & $-$0.269 \\
 Atom.\ stab. & $-$0.305 & $-$0.291\\
  Validity         & $-$0.212 & $-$0.216 \\            \bottomrule                                                                \end{tabular}                                                     
  \end{table}  

Finally, we ablate the choice of timesteps $\mathcal{T}$ at which we measure uncertainty along the generation trajectory. As shown in \Cref{fig:step_profile}, the timesteps closest to the clean sample (small $t$) correlate most negatively with sample quality (atomic stability), supporting our choice to concentrate the uncertainty measurement at the end of the generation process.

\section{Conclusion}
We have introduced a post-hoc uncertainty estimation approach for molecular diffusion models based on noise-prediction variability along the sampling trajectory, and empirically validated its effectiveness on QM9. A central limitation is that the approach does not transfer to the larger, more complex GEOM-Drugs dataset, which we leave to future work. One promising direction is to compute uncertainty in a `semantic' feature space---i.e., based on a pretrained molecular encoder---analogous to the approach shown to be effective for uncertainty estimation in high-dimensional image generation \citep{jazbec2025generative}. Lastly, our uncertainty score is a natural drop-in replacement for the NLL-based verifier in recently proposed adaptive test-time scaling methods \citep{lee2025adaptive}, which represents another promising future direction.

\bibliography{main}
\bibliographystyle{icml2026}

\newpage

\end{document}